\documentclass{article}

\usepackage[preprint,nonatbib]{neurips_2023}
\usepackage{natbib}
\setcitestyle{authoryear}
\usepackage[utf8]{inputenc} % allow utf-8 input
\usepackage[T1]{fontenc}    % use 8-bit T1 fonts
\usepackage{hyperref}       % hyperlinks
\usepackage{url}            % simple URL typesetting
\usepackage{booktabs}       % professional-quality tables
\usepackage{amsfonts}       % blackboard math symbols
\usepackage{nicefrac}       % compact symbols for 1/2, etc.
\usepackage{microtype}      % microtypography
\usepackage[dvipsnames]{xcolor}         % colors
\usepackage{bm}
\usepackage{multirow}
\usepackage{subcaption}
\usepackage{makecell}
\usepackage{graphicx}
\usepackage{algorithmic}
\usepackage{algorithm}
\usepackage{enumitem}
\usepackage{amsmath}
\usepackage{nicematrix}
\usepackage{soul}

\newcommand{\mmonth}{\text{month}}
\newcommand{\llon}{\text{lon}}
\newcommand{\llat}{\text{lat}}

\newcommand{\hlc}[2][yellow]{{%
    \colorlet{foo}{#1}%
    \sethlcolor{foo}\hl{#2}}%
}

\title{Lightweight, Pre-trained Transformers for\\ Remote Sensing Timeseries}

\author{
Gabriel Tseng$^{1,2}$ \quad Ruben Cartuyvels$^{1,3}$ \quad Ivan Zvonkov$^{4}$ \quad Mirali Purohit$^{5}$ \\ \textbf{David Rolnick}$^{1,2}$ \quad \textbf{Hannah Kerner}$^{5}$ \\
$^1$ Mila -- Quebec AI Institute \\ $^2$ McGill University \\ $^3$ KU Leuven \\
$^4$ University of Maryland, College Park \\ $^5$ Arizona State University 
}
\begin{document}

\maketitle

\begin{abstract}
Machine learning methods for satellite data have a range of societally relevant applications, but labels used to train models can be difficult or impossible to acquire. Self-supervision is a natural solution in settings with limited labeled data, but current self-supervised models for satellite data fail to take advantage of the characteristics of that data, including the temporal dimension (which is critical for many applications, such as monitoring crop growth) and availability of data from many complementary sensors (which can significantly improve a model's predictive performance). We present Presto (the \textbf{P}retrained \textbf{Re}mote \textbf{S}ensing \textbf{T}ransf\textbf{o}rmer), a model pre-trained on remote sensing pixel-timeseries data. By designing Presto specifically for remote sensing data, we can create a significantly smaller but performant model. Presto excels at a wide variety of globally distributed remote sensing tasks and performs competitively with much larger models while requiring far less compute. Presto can be used for transfer learning or as a feature extractor for simple models, enabling efficient deployment at scale.
\end{abstract}

\section{Introduction}
\label{sec:intro}

Machine learning is increasingly being applied to the remote sensing domain, in particular to understand the evolution of the Earth's surface over time \citep{DynamicWorld,voosen2020europe,abys2024two,wang2020mapping1}. These applications can have important societally beneficial outcomes, ranging from tracking progress on sustainable development goals \citep{ferreira2020monitoring} to improved weather forecasting \citep{english2013impact, voosen2020europe} to disaster management \citep{kansakar2016review}. However, labeled datasets often contain labels that are few, sparse, and unreliable \citep{bressan2022semantic}, especially for under-resourced geographies, leading to poor global generalization \citep{yifang2015global,rapidresponse, nakalembe2021sowing}. This has spurred the investigation of self-supervised learning algorithms for remote sensing data.

Current self-supervised approaches for remote sensing data have drawn from methods in computer vision, yielding models that treat remote sensing data as single-timestep images \citep{jean2019tile2vec, manas2021seasonal, ayush2021geography}. Such models (i) cannot benefit from patterns that emerge when an area is monitored over time, which is especially important for agriculture and other seasonal landcover, (ii) typically only consider a single satellite product (such as Sentinel-2 multispectral data), despite there being hundreds of publicly available satellite data products \citep{GEEDC}, (iii) are typically large and computationally expensive \citep{reed2022scale,cong2022satmae,fuller2023croma}, making the deployment of these models at scale challenging, and (iv) cannot natively handle the labels for many remote sensing datasets, which are points or irregularly shaped polygons \citep{rao2020sar,batjes2017wosis}, requiring additional methods to handle these labels 
\citep{wang2020weakly}.

We introduce the \textbf{P}retrained \textbf{Re}mote \textbf{S}ensing \textbf{T}ransf\textbf{o}rmer (Presto), a lightweight model designed to ingest pixel-timeseries inputs from a variety of Earth observation sensors and data products. Presto operates on individual pixels, using the temporal and multimodal structure of the data instead of the image structure. To learn powerful representations of remote sensing data that can be adapted to a wide range of tasks, Presto leverages a self-supervised masked autoencoding approach, reconstructing unobserved timepoints and sensory modalities. This allows Presto to be robust to missing data and to flexibly accommodate diverse input formats. We find Presto excels even in image-based tasks where the temporal dimension is completely absent. 

Presto addresses the following requirements, which are critical to the useful deployment of pre-trained models in the remote sensing context:

\begin{figure*}
    \centering
    \includegraphics[width=\linewidth]{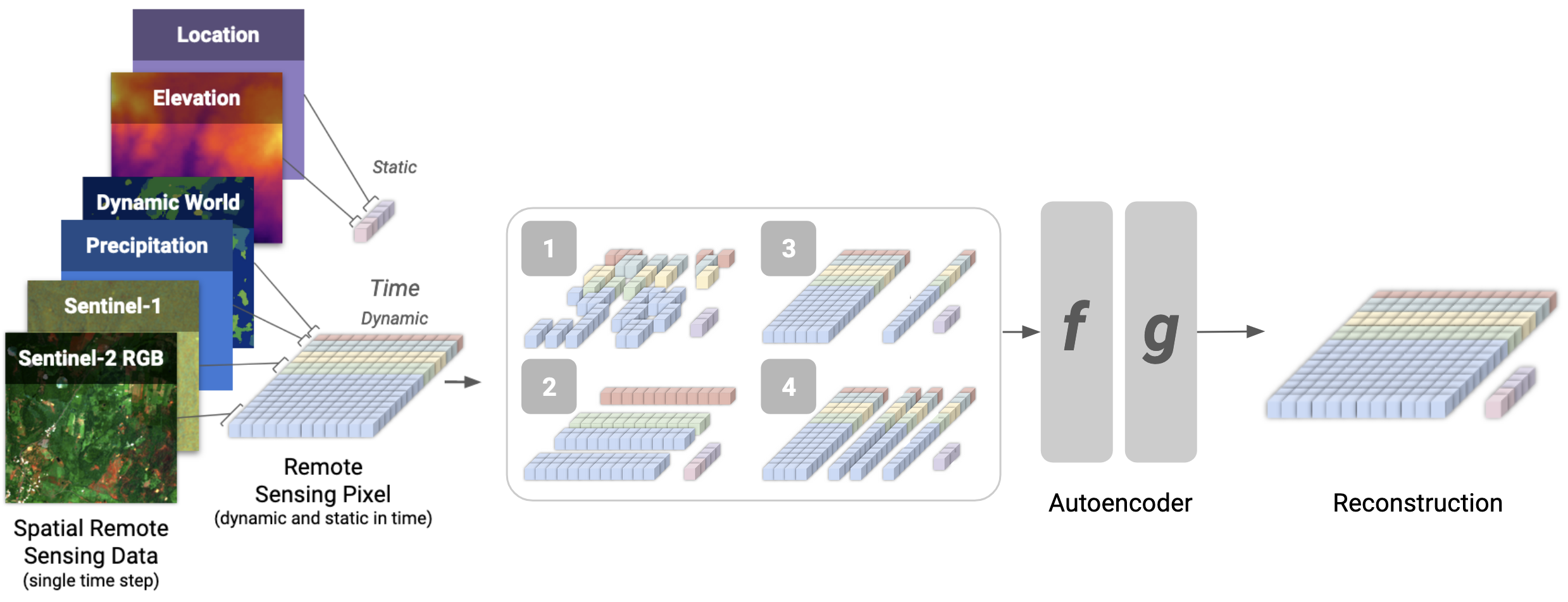}
    \caption{\textbf{Presto learns from structurally-masked remote sensing pixel-timeseries}. We construct a multi-sensor remote sensing pixel-timeseries, and randomly select one of the four masking strategies described in Section \ref{sec:masking}.
    % to mask the data. 
    The encoder-decoder model is trained to reconstruct the original timeseries. At fine-tuning time, we discard the decoder and only use the encoder's output. The downstream task may have incomplete inputs (missing timesteps or sensors) since the encoder is specifically trained on such inputs. Presto receives both static-in-time and dynamic-in-time inputs and the location metadata of each pixel timeseries.}
    \label{fig:training}
\end{figure*}

\begin{itemize}[noitemsep,topsep=0pt,leftmargin=*]
\item \textbf{Computational efficiency}: When deployed, models built for remote sensing data are typically used to make contiguous geospatial predictions over millions (or billions) of samples to form a predicted map. The computational performance of models is therefore one of the primary considerations at deployment time. \citet{worldcerealsbenchmark}, \citet{hengl2017soilgrids250m} and \citet{robinson2019large} are all global- or large- scale map making efforts that prioritized efficiency over accuracy when deploying remote sensing models at scale. Presto is competitive with ViT or ResNet based models, despite having up to $1000 \times$ fewer trainable parameters and requiring orders of magnitude fewer FLOPs at inference time.
\item \textbf{Ability to process inputs of varying shapes}: Different downstream tasks may require very different remote sensing inputs. For example, for crop mapping and yield estimation, \citet{garnot2020psetae} and \citet{You_Li_Low_Lobell_Ermon_2017} discarded all spatial information in the inputs in favor of emphasizing temporal patterns. We test Presto on a wide range of downstream inputs (for example, with spatial information present or absent, and with single or multiple timesteps of data), and find it is competitive with models designed specifically for those inputs.
\item \textbf{Ability to process a range of remote sensing datasets}: For fuel moisture estimation, \citet{rao2020sar} found that the inclusion of derived products in addition to raw inputs significantly improved performance. Presto can ingest a range of static-in-time and dynamic-in-time raw input data as well as derived product inputs widely used in Earth observation (such as NDVI \citep{rouse1974monitoring}). 
\item \textbf{Ability to handle missing data}: The coverage of remote sensing products is often spatially and temporally incomplete. For example, certain regions experience very high ($>90\%$) cloud coverage, reducing the utility of optical measurements such as Sentinel-2 imagery \citep{sudmanns2019assessing}. Because Presto ingests a variety of remote sensing inputs, it can leverage alternative data sources if one is missing (for instance, relying on Sentinel-1, which sees through clouds, if Sentinel-2 images are cloudy). 
\end{itemize}

Our results support the surprising conclusion that a pixel-based approach can in some cases match or outperform sophisticated computer vision-based approaches. We hypothesize that this is possible because (i) Presto learns from many semantically dense data sources, allowing it to extract informative patterns from pixel-timeseries, and (ii) many remote sensing tasks require significantly smaller receptive fields than those provided by computer vision-based models. \citet{DynamicWorld} leveraged such properties to train a model $100\times$ smaller than standard models while achieving state-of-the-art land-cover segmentation results.

\section{Related Work}

\paragraph{Architectures for Remote Sensing} 
When processing remote sensing timeseries, transformers have been extensively investigated either as unmodified architectures \citep{russwurm2020self} or as architectures designed for specific tasks \citep{garnot2020psetae,tarasiou2023vits}. Recurrent networks have also been investigated \citep{rapidresponse,russwurm2020self}. When treating remote sensing data as single or few (up to 3) timestep images, architectures from computer vision are commonly used, ranging from ResNets \citep{manas2021seasonal,ayush2021geography,russwurm2020meta} to Vision Transformers \citep{cong2022satmae,reed2022scale,fuller2023croma}.

\paragraph{Self-supervised learning for Remote Sensing} While contrastive learning has been investigated for remote sensing \citep{manas2021seasonal}, recent self-supervised learning research has focused on masked autoencoders \citep{yuan2022sits,cong2022satmae,reed2022scale,fuller2023croma}. However, these approaches (i) focus on learning from raw satellite data products (ignoring derived products such as elevation) and typically only ingest data from a single sensor (the exception being the CROMA model of \citet{fuller2023croma}, which ingests both Sentinel-1 and Sentinel-2 data), (ii) ingest very few or no timesteps (\citet{reed2022scale} and \citet{fuller2023croma} ingest only one timestep while \citet{cong2022satmae} ingest up to three timesteps), (iii) expect data in a certain size (for instance, ViT based models require spatial dimensions to be present), so that missing data is not handled natively, and (iv) generally yield larger models ranging from 2.5 million parameters \citep{yuan2020self} to over 300 million parameters for ViT-based methods, making their deployment in compute-constrained settings challenging.

\section{Method}  \label{sec:rest}

We aim to learn a model, $f$, which can learn useful representations in a self-supervised manner given unlabelled remote sensing pixel-timeseries data while meeting the usability requirements outlined in Section \ref{sec:intro}. This model can then be applied to a wide variety of downstream remote sensing tasks. These downstream tasks may contain input data from a range of sensors with differing numbers of timesteps.

Our approach is based on the masked autoencoding framework \citep{he2022masked}, in which the network architecture includes both an encoder ($f$) and a decoder ($g$). During pre-training, part of the input is masked out and the encoder embeds the remaining (non-masked) part of the input. The decoder aims to reconstruct the masked-out part of the input, given the encoder's output. At fine-tuning time, we discard $g$ and only use $f$ (either as a feature extractor or a fine-tuneable model) for downstream tasks. In the sections below, we discuss how Presto customizes this general framework for multi-sensor remote sensing timeseries data. An overview of the Presto pre-training methodology is shown in Figure \ref{fig:training}, and full pre-training details are in Section \ref{app:pretraining}.

\subsection{Pre-training Data} \label{sec:data}

Self-supervised models for remote sensing must generalize to a wide range of geographies and tasks \citep{lacoste2023geo}. We therefore aimed to collect a globally representative pre-training dataset. We followed the sampling strategy of \citet{DynamicWorld} to construct a dataset of 21.5M pixel samples, each with a resolution of 10m per pixel. Appendix \ref{app:data} describes the pre-training dataset construction process in detail.
Presto was trained on pixel-timeseries of 12-month contiguous intervals, sampled from a 2-year period from the beginning of 2020 until the end of 2021, with each month represented by one timestep (similar to the approach adopted by \citet{tseng2021cropharvest}). 
Derived data products that result from the analysis of lower level data (e.g., \citet{parkinson2006earth}) can significantly improve model performance \citep{rao2020sar, hengl2017soilgrids250m}. We therefore pre-trained Presto on a diverse set of directly-sensed and derived Earth observation products which we pre-processed and exported using Google Earth Engine \citep{gorelick2017google}.

A pre-training batch contained several pixel-timeseries samples, each of which is a concatenation of dynamic-in-time datapoints with each timestep representing a month (yielding $T=12$ timesteps in total). The following dynamic-in-time data products were used, yielding $15$ channels: (i) Sentinel-2 (S2) multispectral data, (ii) Sentinel-1 (S1) radar data, (iii) ERA5 climate reanalysis data, (iv) NDVI \citep{rouse1974monitoring} derived from Sentinel-2 data and (v) land cover classes $\mathcal{V}$ from Dynamic World. To every pixel-timeseries we appended two static-in-time products: (i) topography data from the SRTM digital elevation model \citep{srtm2003cgiar} and (ii) location coordinates of each pixel. Hence, one pre-training sample $x$, comprising a pixel-timeseries $t \in [\mathbb{R}^{T \times 15}; \mathcal{V}^{T \times 1}]$ and static variables $s \in \mathbb{R}^{1 \times 5}$, is summarized as follows: 
\begin{equation}
x = \Big[ \big\{ t_i^\text{S1};\ t_i^\text{S2} ;\ t_i^\text{ERA5};\ t_i^\text{NDVI};\ t_i^\text{DW} \ | \ i = 1, ..., 12 \big\}; \ s^\text{TG}; \ s^\text{Loc} \Big]
\end{equation}
From now on, we use ``pixel-timeseries'' to refer to both the dynamic and the static variables.

\subsection{Encoding and tokenization} \label{sec:tokenization}

\begin{figure}
    \centering
    \includegraphics[width=\linewidth]{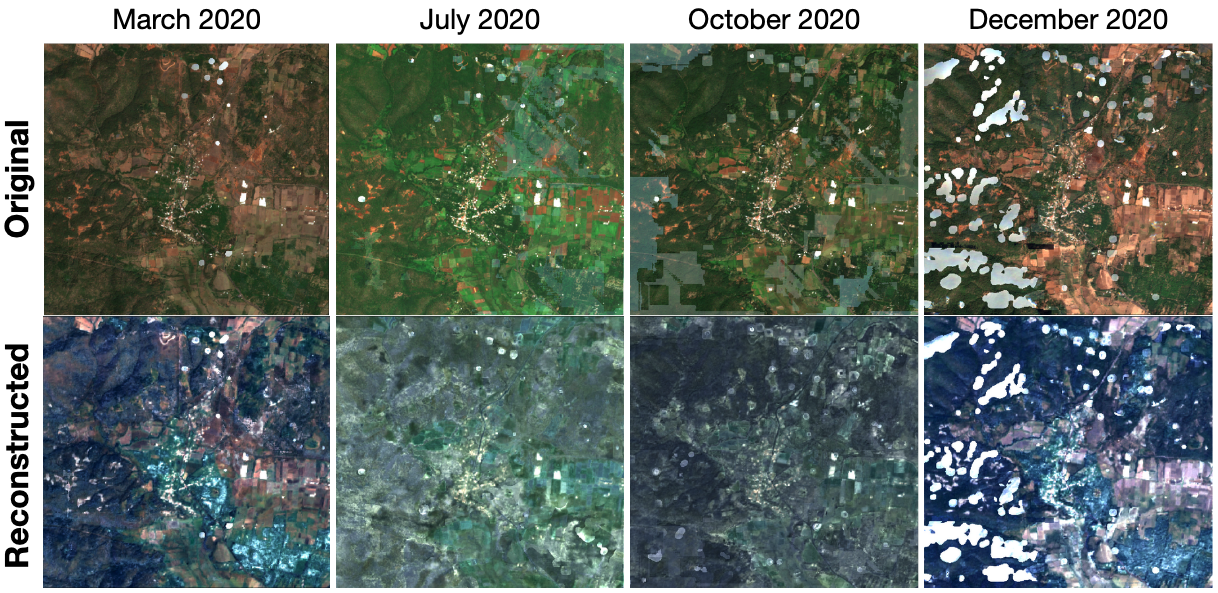}
    \caption{\textbf{Presto learns to reconstruct channels that are completely masked in a spatially cohesive manner}. In this experiment, we masked only the Sentinel-2 RGB channels; Presto was able to reconstruct these channels even when they were absent from the input. The reconstructions are spatially consistent even though Presto only receives single pixel inputs.}
    \label{fig:reconstruction}
\end{figure}

We transformed the pixel-timeseries $x$ into a number of tokens (each represented by an embedding $e$) to be processed by the Presto transformer. 
Per timestep $0\le i < T$, we split the input variables into channel groups $\mathcal{C}$ according to their type of sensor or source: e.g., the S1 bands form one channel group. We describe these groups in more detail in Appendix \ref{sec:channel_groups}.
Each real-valued channel group represents a different sensor, native spatial resolution or (in the case of Sentinel-2 channel-groups) region of the electromagnetic spectrum. We projected each channel group to a common latent space of dimension $d_e$ by separate learned linear projections $h^\mathcal{C}$: e.g., $e_i^\text{S1} = h^\text{S1}(t_i^\text{S1})$.
The Dynamic World classes are categorical, so we embedded them by indexing them into an embedding matrix.

\begin{table}[]
    \centering
    \caption{\textbf{We evaluated Presto on a wide variety of downstream tasks}, including segmentation (seg.), multi-label (ml) scene classification (class.) and regression (reg.) tasks. There is diversity in terms of data composition, geographic area and training set size. Input shape describes the shape of a single sample, in terms of [Height, Width, Timesteps, Channels]. We \textbf{bold} the temporal dimension, to highlight time-series versus single-timestep inputs.}
    \label{tab:eval_sets}
    \vskip 0.15in
    \begin{NiceTabular}{llrrr}
    \CodeBefore
    \rowcolor{gray!20}{2,3,4,6,7,10}
    \Body
    \toprule[1.5pt]
        Dataset & Task & Region & \makecell{Input shape \\ [H, W, \textbf{T}, C]} & 
        \makecell{Train 
        \\samples} \\
        \toprule[1.5pt]
         \multirow{3}*{CropHarvest} & \multirow{3}*{Seg.} & Kenya & \multirow{3}*{[1, 1, \textbf{12}, 18]} & 1,345 \\
         & &  Brazil & & 203 \\
         & & Togo & & 1,319 \\
        S2-Agri$_{100}$ & Class. & France & [5, 5, \textbf{24}, 10] & 1,500 \\
        \multirow{2}*{TreeSat} & \multirow{2}*{\makecell[l]{ML \\ Class.}} & \multirow{2}*{Germany} & [6, 6, \textbf{1}, 2] & \multirow{2}*{45,337} \\
        & & & [6, 6, \textbf{1}, 11] & \\
        \multirow{2}*{EuroSat} & \multirow{2}*{Class.} & \multirow{2}*{Europe} & [64, 64, \textbf{1}, 3] & \multirow{2}*{21,600} \\
        & & & [64, 64, \textbf{1}, 11] & \\
        Fuel Moisture & Reg. & USA & [1, 1, \textbf{3}, 19] & 1,578 \\
        Algae Blooms & Reg. & USA & [1, 1, \textbf{12}, 19] & 777 \\
        \bottomrule[1.5pt]
    \end{NiceTabular}
\end{table}

Unlike natural images in which the data and its label are self-contained, remote sensing labels are inherently associated to a place and time on Earth (i.e., a latitude/longitude and timestamp). In addition, while natural images contain RGB channels from the same camera sensor, Presto's pixel-timeseries input contains channels from multiple remote sensing instruments and data products. We therefore wanted to communicate to the model: (i) the location of the datapoint (already present in the input as static variable through coordinates $s_\text{Loc}$) and a variable's (ii) timestamp and (iii) channel group. 
We did this by adding encodings to the previously described embeddings $e$. The complete encoding has dimension $d_e$ and contains a concatenation of positional, month, and learned channel encodings described below.

\begin{itemize}[topsep=0pt,leftmargin=*]
    \item \textbf{Positional:} We used the sinusoidal positional encoding originally used by \citet{vaswani2017attention}.
\item \textbf{Month:} We added an encoding representing the month being captured by each token, because we expect timesteps from similar months to have similar features even if they are from different years. We assign an integer to each month ranging from $0$ to $11$, yielding:
\begin{align}
    p_{\mmonth, 2i} &= \sin\left((2\pi \times \mmonth) / 12\right) \\ p_{\mmonth, 2i+1} &= \cos\left((2\pi \times \mmonth) / 12\right)
\end{align}
For static-in-time variables, the positional and month encodings were set to zero.
\item \textbf{Channel Group:} 
Each token is associated with a set of input channels. In multispectral SatMAE \citep{cong2022satmae}, a fixed encoding was used to communicate input-band information with different channels representing different wavelengths, which is possible because only input data from one sensor (Sentinel-2) is used. 
However, since Presto's input data includes multiple remote sensing products, we applied a learnable encoding for each channel group from the set of possible channel groups $\mathcal{C} = \{ \text{S1}, \text{S2 RGB}, ..., \text{ERA5}, \text{TG}, \text{Loc} \}$. 
\end{itemize}

The transformer input $E \in \mathbb{R}^{(T \cdot |\mathcal{C}_{\textrm{dynamic}}| + |\mathcal{C}_{\textrm{static}}|) \times d_e}$ (for encoder dimension $d_e$) is a concatenation of:
\begin{itemize}[noitemsep,topsep=0pt,leftmargin=*]
\item Dynamic variables, for timesteps $i < T$ and channel groups $c \in \mathcal{C}$: $e^c_i = h^\text{c}(t_i^\text{c}) + [ p_\text{channel}^\text{c}; \ p_\text{sin(i)}; \ p_\text{month(i)}]$
\item Topographical data: $e^\text{TG} = h^\text{TG}(s^\text{TG}) + [p_\text{channel}^\text{TG}; \ 0 ; \ 0]$
\item Coordinates: $e^\text{Loc} = h^\text{Loc}(s^\text{Loc})$
\end{itemize}

\begin{table}
    \centering
    \footnotesize
    \caption{Mean F1 score across all CropHarvest tasks. Presto outpeforms TIML \citep{tseng2021timl} and MOSAIKS-1D \textbf{while requiring the adaptation of far fewer parameters.} The TIML and MOSAIKS-1D model did not receive Dynamic World as input, so we measured Presto's performance both with and without it.}
    \label{tab:cropharvest-results}
    \vskip 0.15in
  \begin{NiceTabular}{lrrr}
\CodeBefore
    \rowcolor{MidnightBlue!20}{6,7}
    \Body
    \toprule[1.5pt]
    & \multicolumn{2}{c}{\#. parameters} & \\
     Model & Total & Adapted & Mean F1\\
     \toprule[1.5pt]
     Random Forest & & & 0.441 \\
     MOSAIKS-1D$_{R}$ & 418K & 8193 & 0.738 \\
      TIML & 91K & 91K & $0.802$ \\
      \midrule
      Presto$_{R}$ & \multirow{2}*{402K} & \multirow{2}*{129} & $0.835$ \\
      \hspace{3mm} no DW & & & $\bm{0.836}$\\
    \bottomrule[1.5pt]
  \end{NiceTabular}
\end{table}

\subsection{Pre-training via Structured Masking} \label{sec:masking}

A key requirement for Presto was to perform well even with incomplete inputs (i.e., when there are missing timesteps, channels, or both). When masking out part of the input $x$, we therefore tailored the masking strategies to encourage the model to learn representations that perform well when given a subset of bands or timesteps for downstream tasks. 
For a $T \times D$ input of $T$ timesteps and $D$ total input channels, we used the following masking techniques (illustrated in Figure \ref{fig:training}), where Presto considers a token to be a $1 \times d$ input (a single timestep of $d$ grouped channels). The coordinates were never masked but the static topological tokens can be.
\begin{enumerate}[noitemsep,topsep=0pt,leftmargin=*]
\item \textbf{Random}: $(t \times d)$ masked values, with $t < T$ and $d < D$
\item \textbf{Channel-groups}: $(T \times d)$ masked values, with $d < D$
\item \textbf{Contiguous timesteps}: $(t \times D)$ masked values, $t < T$
\item \textbf{Timesteps}: $(t \times D)$ masked values, with $t < T$
\end{enumerate}
For each training instance, we randomly sampled from the above strategies to construct a mask. 

To handle both the categorical and continuous inputs we used the following loss function, which balances the continuous and categorical losses for every batch so that each reconstructed value receives the same weighting in the final loss:
    $\mathcal{L}_\text{total} = \mathcal{L}_\text{MSE} + \lambda \frac{N_\text{cat}}{N_\text{cont}}\mathcal{L}_\text{CE}$.
$\mathcal{L}_\text{MSE}$ is the mean squared error reconstruction loss used for the continuous values, $\mathcal{L}_\text{CE}$ is the cross entropy loss used for the categorical values, $N_\text{cont}$ is the number of masked continuous values and $N_\text{cat}$ is the number of masked categorical values in the batch. $\lambda$ is a hyperparameter, which we set to $2$.

\begin{figure}
\centering
\includegraphics[width=0.7\linewidth]{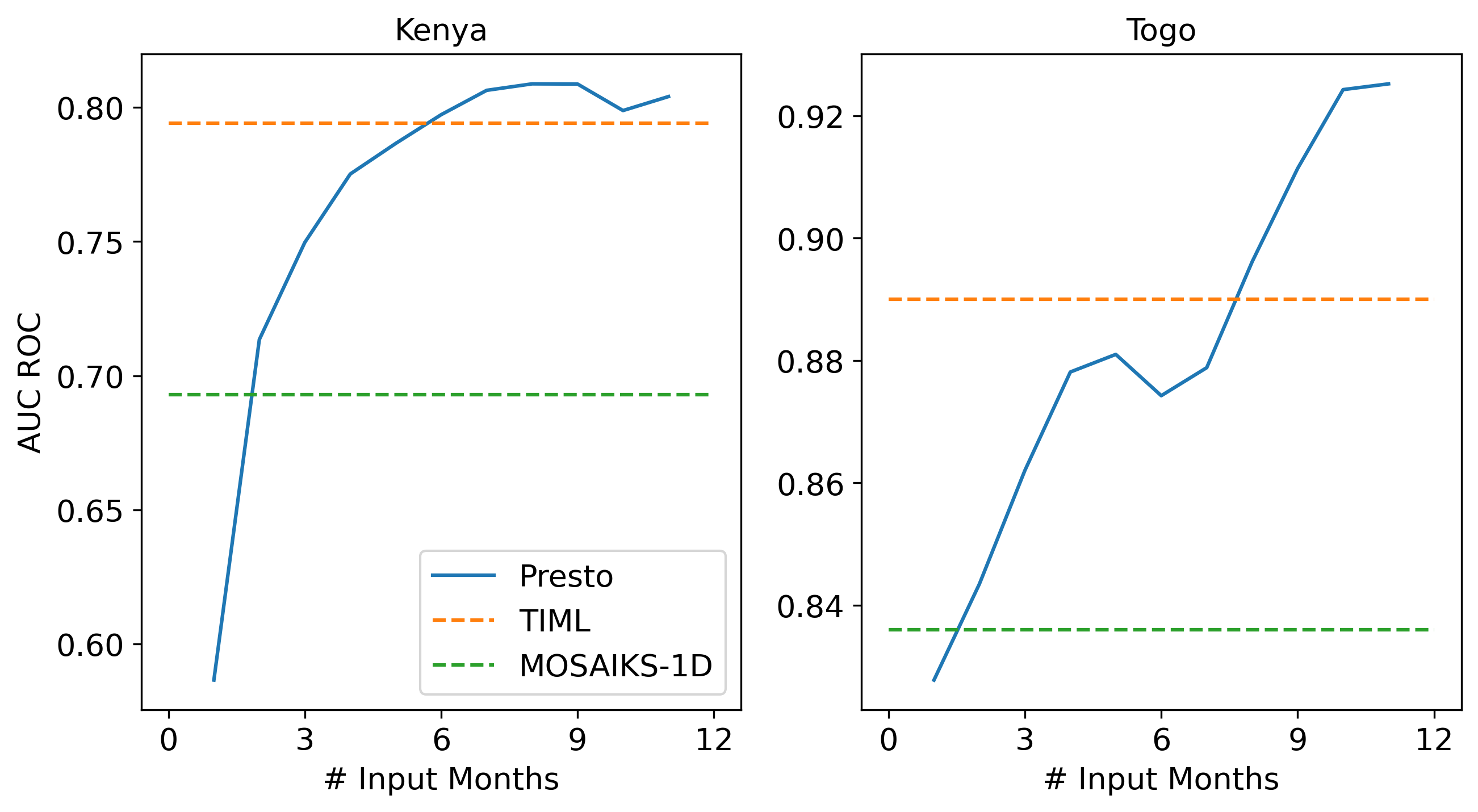}
\caption{\textbf{Presto is robust to incomplete inputs}. We measured the AUC ROC score of Presto with Linear probing (Presto$_{R}$) on the CropHarvest dataset when no Dynamic World input is passed, and with a subset of input months (the x-axis). We plot the performance of MOSAIKS-1D and TIML when they receive the full 12 months of input (dashed horizontal lines) - Presto$_{R}$ recovered the performance of these models given only a subset of input months.} \label{fig:cropharvest_over_time}
\end{figure}

\section{Experiments} \label{sec:experiments}
In all experiments described below, we use a Presto model with identical encoder and decoder configurations (2 attention layers with 8 heads, an embedding size of 128 and an MLP ratio of 4). We investigated the effect of different encoder configurations in Table \ref{tab:scaling}. 

For downstream evaluation, we took the encoder-decoder model learned during pre-training and discarded the decoder. As in \citet{he2022masked}, we passed a global pool of all the encoder's output tokens to a downstream classifier. We evaluated the performance of three different models: Presto$_{R}$, Presto$_{RF}$, and Presto$_{FT}$, defined below.
\begin{itemize}[noitemsep,topsep=0pt,leftmargin=*]
    \item \textbf{Feature extraction.} \citet{rolf2021generalizable} demonstrated the utility of neural networks as feature-extractors on top of which computationally efficient classifiers could be trained. Presto$_{R}$ and Presto$_{RF}$ consist respectively of linear or logistic regressions and random forests trained on Presto's embeddings. Since only the regression/random forest is trained, this a computationally efficient method for adapting Presto to a wide range of tasks.
    \item \textbf{Fine-tuning}. Presto$_{FT}$ consists of the Presto encoder, followed by a linear transformation of the pooled tokens to the desired outputs. This entire model (the encoder and the linear transformation) is fine-tuned on the training data from each evaluation task. We used a subset of the (downstream) training data for validation.
\end{itemize}

During pre-training, we used a \textbf{validation task} consisting of classifying all points in the CropHarvest dataset \citep{tseng2021cropharvest} according to their FAO indicative crop classifications. For this validation task, we excluded points used for evaluation (Section \ref{sec:timeseries}).

For evaluation, we compared Presto to state-of-the-art task-specific baselines (Section \ref{sec:eval}). Because there are no other global self-supervised models for pixel-timeseries, we adapted MOSAIKS \citep{rolf2021generalizable} for timeseries data by performing convolutions over the temporal rather than spatial dimension (MOSAIKS-1D). We used the output features with random forests (MOSAIKS-1D$_{RF}$) and regressions (MOSAIKS-1D$_{R}$).

\begin{figure}\centering\includegraphics[width=\linewidth]{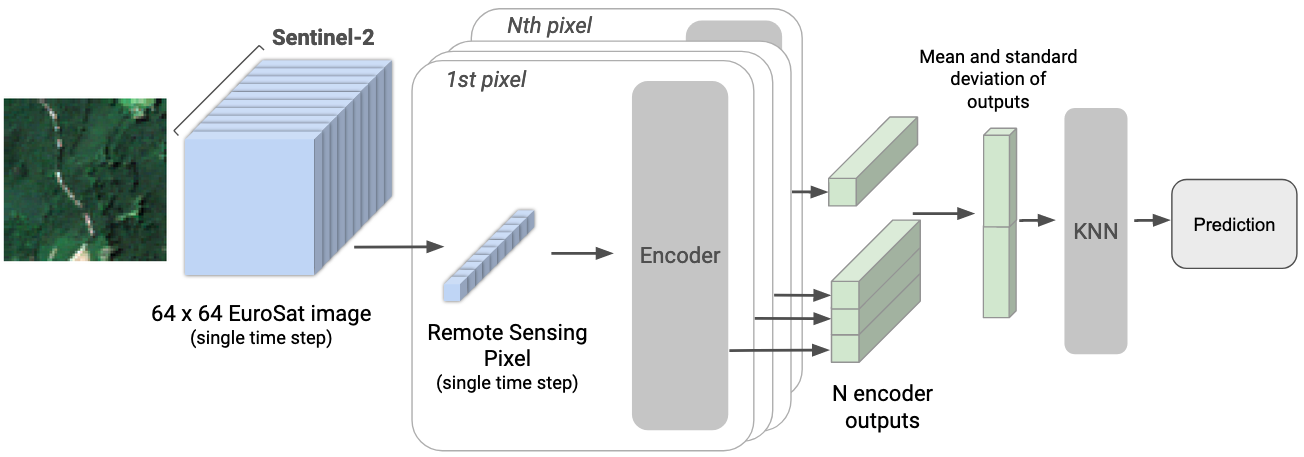}
    \caption{We obtained per-image predictions using Presto by computing a mean and standard deviation of Presto's per-pixel outputs, and passing this concatenated vector to a downstream classifier. We illustrate this for the EuroSat task.}
    \label{im:aggregates}
\end{figure}

\section{Evaluation Tasks \& Results}\label{sec:eval}
We evaluated Presto using six evaluation tasks spanning diverse task types, geographic locations (4 continents and 38 countries), input data modalities, and fine-tuning dataset sizes (Table \ref{tab:eval_sets}). Whenever possible, we benchmarked Presto against the state-of-the-art model for that task.

\textbf{Applying Presto to downstream tasks is computationally efficient}. While other methods require a cluster of GPUs for fine-tuning \citep{cong2022satmae}, we fine-tuned Presto on a single GPU or CPU. For the fuel moisture task described in Section \ref{sec:timeseries}, fine-tuning Presto took under 6 minutes on a 2017 MacBook Pro's CPU. When Presto is used as a feature extractor, simple models can be trained which require few parameters to be learned, as we show in Table \ref{tab:cropharvest-results}. Even when fully fine-tuned, Presto's small size meant that relatively few parameters needed to be trained (Tables \ref{tab:eurosat_finetune} and \ref{tab:s2agri}). This makes Presto accessible to practitioners, especially those lacking significant computational resources. 

Below, we describe the tasks used to evaluate Presto and discuss Presto's performance on these tasks.

\begin{table}
    \centering
    \footnotesize
    \caption{RMSE results on the regression tasks. The literature baselines are not directly comparable, since they use different input datasets or private test data (or both). \citet{rao2020sar} reported an RMSE of 25 on the fuel moisture dataset with a physics-assisted neural network and the algae bloom competition winner reported an RMSE of 0.761, indicating our results are within the scope of utility. Best results are {\color{blue} \textbf{highlighted blue}}, with second best results in \textbf{bold}. Models have a high variance in performance across tasks, so we calculated the mean difference in RMSE from the linear regression baseline across both tasks. Presto performed most consistently, both when used as a feature-extractor and when fine-tuned.}\label{tab:reg}
    \vskip 0.15in
    \begin{NiceTabular}{lrrr}
    \CodeBefore
    \rowcolor{orange!20}{5}
    \rowcolor{MidnightBlue!20}{6,7}
    \Body
    \toprule[1.5pt]
    & \makecell{Fuel \\ Moisture} & \makecell{Algae \\ Blooms} & \makecell{Mean \\ difference} \\
    \toprule[1.5pt]
    Linear Regression & 28.20 & \bm{$0.850$} & 0\% \\
    Random Forest & {\color{blue} \bm{$23.84$}} & $1.249 $ & 15.7\%\\
    MOSAIKS-1D$_{RF}$ & $28.75$ & $0.972$ & 8.15\%\\
    Presto$_{FT}$ (random init.) & $26.07$ & $0.955$ & 2.40\% \\
    \midrule[0.5pt]
    Presto$_{FT}$ & \bm{$25.28$}  & {\color{blue}\bm{$0.815$}} & ${\color{blue}\bm{-7.24\%}}$ \\
    Presto$_{RF}$ & $25.98$ & $0.884$ & \bm{$-1.94\%$} \\
    \bottomrule[1.5pt]
    \end{NiceTabular}
\end{table}

\subsection{Timeseries Tasks}\label{sec:timeseries}
\begin{itemize}[noitemsep,topsep=0pt,leftmargin=*]
\item \textbf{Crop type Segmentation}: The CropHarvest \citep{tseng2021cropharvest} evaluation datasets consist of binary pixel classification of (i) maize in Kenya, (ii) coffee in Brazil and (iii) cropland in Togo. We compared Presto to the baselines provided by CropHarvest and to Task-Informed Meta-Learning \citep[TIML,][]{tseng2021timl}, which achieved state-of-the-art results on these datasets.
\item \textbf{Fuel Moisture}: The live fuel moisture dataset \citep{rao2020sar} measures live fuel moisture content in the Western U.S. \citet{rao2020sar}'s baseline used 5-fold cross validation to evaluate model performance; for future comparability, we used a single geographically separated test set (a test set covering a different geographic area than the training set).
\item \textbf{Algae Blooms}: The algae blooms dataset \citep{algaeblooms} measures the severity of cyanobacterial algal blooms in different parts of the U.S. We used the subset of the dataset in the Midwestern U.S. The dataset was originally released as part of a competition, so the test data is not available. In addition, competitors could download many Earth observation datasets to train their models, making direct comparisons to competition results difficult. Since the competition's winning solution used a tree-based method, we benchmarked against a regression and a random forest using a geographically separated test set.
\end{itemize}

\begin{table}
\centering
\footnotesize
\caption{Results on the TreeSatAI dataset. We compared Presto to the dataset's benchmark models. The MLPs contain 3 layers (with 563K-723K parameters respectively) and are tuned for this task. We froze the Presto encoder's 402k parameters and trained a random forest on its outputs with default scikit-learn hyperparameters.}
    \label{tab:treesat}
    \vskip 0.15in
\begin{NiceTabular}{llrrrr}
    \CodeBefore
    \rowcolor{MidnightBlue!20}{5,8}
    \Body
\toprule[1.5pt]
& & \multicolumn{2}{c}{Weighted} & \multicolumn{2}{c}{Micro} \\
Model & Data & $F_1$ & mAP & $F_1$ & mAP \\
\toprule[1.5pt]
MLP & \multirow{3}*{S1} & 10.09 & 29.42 & 12.82 & 33.09 \\
LightGBM & & 11.86 & 32.79 & 14.07 & 35.11 \\
Presto$_{RF}$ & & $\bm{38.34}$ & $\bm{35.45}$ & $\bm{40.79}$ & $\bm{38.64}$ \\
\midrule[0.5pt]
MLP  & \multirow{3}*{S2} & $51.97$ & \bm{$64.19$} &$54.59$ & \bm{$65.83$}\\
LightGBM & & 48.17 & 61.99 & 52.52 & 61.66 \\
Presto$_{RF}$ &  & $\bm{55.29}$ & 61.53 & $\bm{58.29}$ & 63.31 \\
\bottomrule[1.5pt]
\end{NiceTabular}
\end{table}

\subsubsection{Timeseries Results}
Presto excels at timeseries tasks, significantly outperforming the state-of-the-art for CropHarvest (Table \ref{tab:cropharvest-results}) and outperforming all baselines for the regression tasks (Table \ref{tab:reg}). 

We found that \textbf{Presto is performant when passed only a subset of timesteps} compared to the 12 timesteps used for pre-training. Presto remained performant when receiving only 3 input timesteps for the fuel moisture task (Table \ref{tab:reg}). We also evaluated Presto when a subset of input months are passed for the CropHarvest dataset (Figure \ref{fig:cropharvest_over_time}). Using a subset of the 12 months, Presto surpassed the performance of TIML and MOSAIKS-1D which used all input months.

Presto is also robust to the removal of input channels. On the CropHarvest dataset (Table \ref{tab:cropharvest-results}), Presto remained performant without the Dynamic World input, showing a negligible difference in mean F1 score compared to the full input.

\subsection{Image-based Tasks} \label{sec:image}

Presto is designed to ingest single pixel-timeseries. When one prediction is required for a set of pixels (as  for image-based tasks and the Image-Timeseries tasks in Section \ref{sec:image_and_time}), we used the following approach to obtain per-image predictions from Presto’s pixel outputs (Figure \ref{im:aggregates}): (i) we encoded the pixels in an image individually, yielding N output tokens, (ii) we calculated the mean and standard deviation of these N output tokens per dimension and concatenated the result, yielding a $2d_e$-dimensional vector (where $d_e$ is Presto’s output token size, or 128), and (iii) we passed this mean and standard deviation vector to a downstream classifier.

\begin{itemize}[noitemsep,topsep=0pt,leftmargin=*]
\item \textbf{TreeSatAI}: The TreeSatAI dataset consists of detecting the presence of one or more tree species (out of 20 possible species) in forestry images in Germany \citep{ahlswede2023treesatai}. We used the train and test splits provided by \citet{ahlswede2023treesatai} and compared Presto to the deep learning and tree-based baselines provided with the dataset. As done for the baselines, we evaluated models using only Sentinel-2 (S2) or Sentinel-1 (S1) data.
\item \textbf{EuroSAT}: The EuroSAT dataset classifies Sentinel-2 multispectral images in Europe with one of 10 landcover classes \citep{helber2019eurosat}. We used the train and test splits provided by \citet{neumann2019domain}. We compared Presto to SatMAE, ConvMAE and ScaleMAE using a k Nearest Neighbors (kNN) classifier at a variety of input resolutions, as was done by \citet{reed2022scale}. We also compared fine-tuned Presto against Seasonal Contrast (SeCo) \citep{manas2021seasonal} and Geography-Aware Self-Supervised Learning (GASSL) \citep{ayush2021geography}. EuroSAT provides all multispectral Sentinel-2 bands, but most other models ingest only RGB images. We evaluated Presto both when it received all multispectral bands as input (MS) and when it only received the RGB bands.
\end{itemize}

\begin{figure}
\centering
\includegraphics[width=0.5\linewidth]{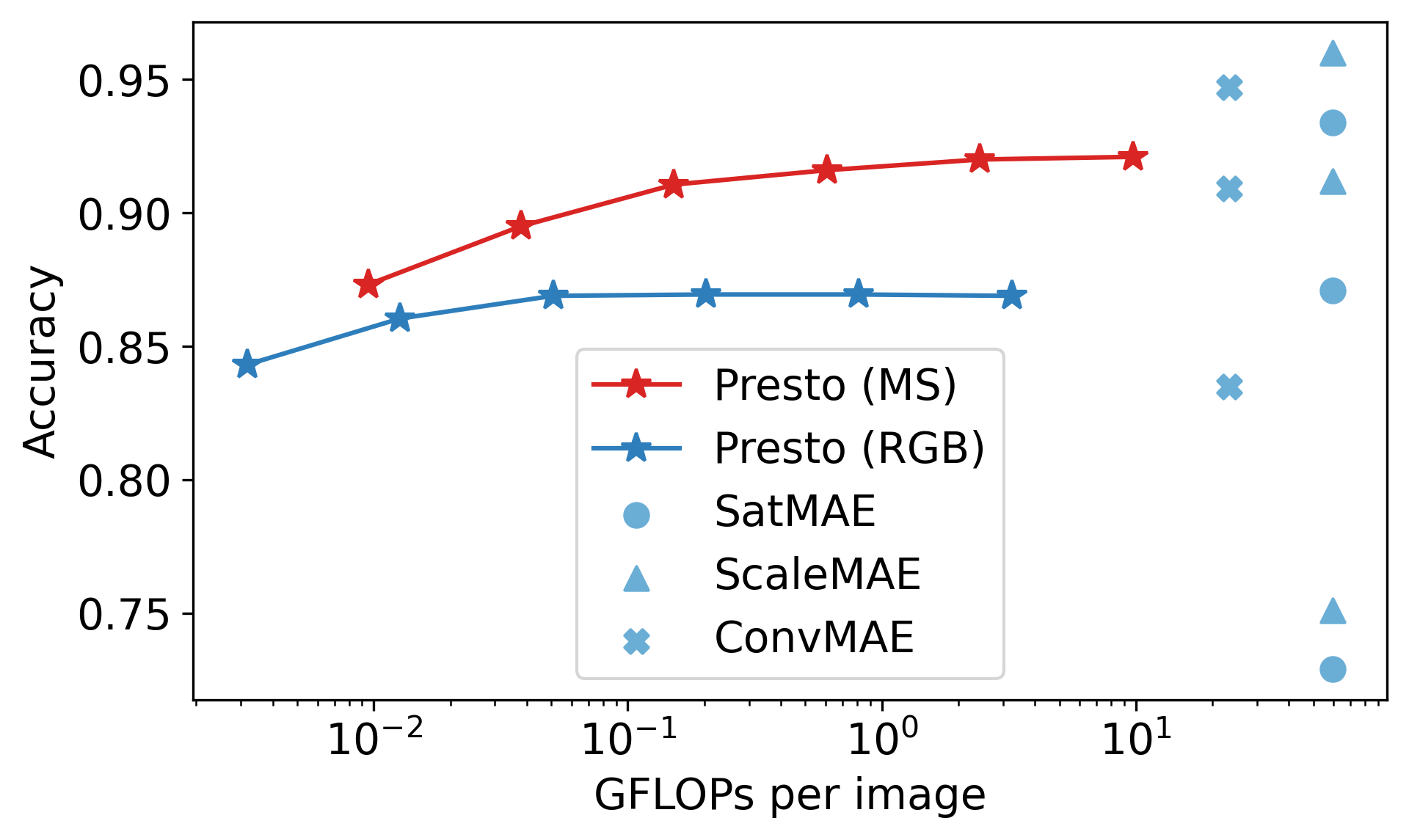}
 \caption{EuroSat accuracy of a kNN@5 classifier given pre-trained model embeddings at a variety of input resolutions (following \citet{reed2022scale}) as a function of FLOPs required to encode an image (note the log scale on the x-axes). All image-based models resized images to $224\times 224$, so the FLOPs required to encode an image do not change with image resolution. Presto achieved competitive results with image-based models while requiring up to four orders of magnitude less FLOPs to encode an image.
 }
 \label{results:eurosat}
\end{figure}

\subsubsection{Image-based Results}
Despite being pre-trained on pixel-timeseries data, \textbf{Presto is competitive on single-timestep image datasets against much larger models}. We followed the setup of \citet{reed2022scale} in measuring the performance of a kNN-classifier on Presto's output embeddings for the EuroSat dataset at varying resolutions. Presto achieved comparable average accuracy (over all image resolutions) to larger ViT-based models with RGB data and significantly outperformed these models with multispectral (MS) data (Figure \ref{results:eurosat}), while requiring orders of magnitude less compute to encode the images in both cases and for any resolution.

\textbf{Presto is performant even when only a small subset of input channels are available} compared to the pre-training channels. For the EuroSAT task (Table \ref{results:eurosat}), Presto received either the full Sentinel-2 input or only RGB bands (which represent only a single token, since only one timestep is available). Similarly, we evaluated Presto when it receives either Sentinel-2 or Sentinel-1 data for the TreeSatAI task (Table \ref{tab:treesat}). 
In both cases, Presto was competitive with methods designed to ingest single-timestep, single-sensor data.

\begin{table}
\centering
\footnotesize
\caption{EuroSAT finetuning accuracy. Presto is the only backbone that can handle both MS and RGB inputs (separate SatMAE models are trained for RGB and MS inputs). We reported Presto results for full resolution; results at reduced resolutions are in Table \ref{tab:eurosat_finetune_full}.} \label{tab:eurosat_finetune}
\vskip 0.15in
\begin{NiceTabular}{lllrr}
    \CodeBefore
    \rowcolor{orange!20}{6,7}
    \rowcolor{MidnightBlue!20}{8,9}
    \Body
\toprule[1.5pt]
     & Backbone & Inputs & \makecell{Params \\ (M)} & Accuracy \\
    \toprule[1.5pt]
    GASSL & ResNet-18 & RGB & 11.69 & 0.895 \\
    SeCo & ResNet-18 & RGB & 11.69 & 0.931 \\
    SatMAE & ViT-Large & RGB & 303.10 & 0.955 \\
    SatMAE & ViT-Large & MS & 305.96 & 0.990 \\
    \midrule
    \multirow{2}*{\makecell{Random \\ init.}} & \multirow{2}*{Presto} & RGB & \multirow{2}*{0.40} & 0.745 \\
    &  & MS &  & 0.924 \\
    \midrule
    \multirow{2}*{Presto} & \multirow{2}*{Presto} & RGB & \multirow{2}*{0.40} & 0.849 \\
    &  & MS &  & 0.953 \\
    \bottomrule[1.5pt]
\end{NiceTabular}
\end{table}

\subsection{Image-Timeseries Tasks} \label{sec:image_and_time}
\begin{itemize}[noitemsep,topsep=0pt,leftmargin=*]
\item \textbf{S2-Agri$_{100}$}: The S2-Agri dataset \citep{garnot2020psetae} classifies crop types in agricultural parcels. We used a variant of S2-Agri (S2-Agri$_{100}$) developed by \citet{yuan2022sits} for the SITS-Former model in which 100 parcels for each crop type are used for training and validation respectively (all other parcels are used for testing), and a $5 \times 5$ pixel patch from each parcel is used for input. We benchmarked Presto against both the pre-trained and randomly initialized SITS-Former model.
\end{itemize}

\subsubsection{Image-Timeseries Results}
The S2-Agri$_{100}$ dataset consists of 24 timesteps at 10 to 30 day intervals (compared to Presto's pre-training data, which consists of 12-month timeseries). Presto remained performant on this dataset, achieving comparable results with SITS-Former despite having $6\times$ fewer parameters (shown in Table \ref{tab:s2agri}). This shows that \textbf{Presto can ingest timeseries at different temporal resolutions and at varying intervals}.

In addition, the S2-Agri dataset is missing pixel location metadata, which is always passed to Presto during pre-training. S2-Agri was sampled from a single S2-tile, so we used the location of the central pixel of this tile for all pixels in the dataset. Even with this much less accurate location metadata, Presto remained performant.

\begin{table}
\centering
\footnotesize
\caption{Results on the S2-Agri$_{100}$ dataset. We followd \citep{yuan2022sits} in reporting overall accuracy (OA), Kappa Cohen score ($\kappa$) and macro-F$_1$ score. All results are an average of 3 runs - standard errors are reported in Table \ref{tab:s2agri-full}.} \label{tab:s2agri}
\vskip 0.15in
\begin{NiceTabular}{lccrrrr}
    \CodeBefore
    \rowlistcolors{4}{orange!20}[cols={3-6}]
    \rowlistcolors{5}{MidnightBlue!20}[cols={3-6}]
    \Body
\toprule[1.5pt]
     & \makecell{Params \\ (M)} & \makecell{Pre \\ Trained?} & OA & $\kappa$ &$F_{1}$ \\
    \toprule[1.5pt]
    \multirow{2}{*}{\makecell{SITS \\ Former}} & \multirow{2}{*}{2.5} & & 65.13 & 0.55 & 42.12 \\
    & & \checkmark & 67.03 & 0.56 & $\bm{42.83}$ \\
    \midrule
    \multirow{2}{*}{Presto} & \multirow{2}{*}{0.4} & & 45.98 & 0.35 & 27.45 \\
    & & \checkmark & $\bm{68.89}$ & $\bm{0.58}$ & 40.41\\
    \bottomrule[1.5pt]
\end{NiceTabular}
\end{table}

\subsection{Ablations}

We conducted three ablations to better understand Presto's performance:

\begin{table}
\centering
\caption{\textbf{Structured masking strategies yield the best downstream performance}. We measured Presto$_{R}$'s F1 score on the CropHarvest validation task. Combining structured strategies outperformed the ``Random'' masking employed by \citep{he2022masked}.} \label{tab:mask_ablation}
\vskip 0.15in
\footnotesize
\begin{NiceTabular}{ccccr}
    \CodeBefore
    \rowcolor{MidnightBlue!20}{6}
    \Body
\toprule[1.5pt]
\makecell{Channel \\ Groups} &  Random & Timesteps & \makecell{Contiguous \\ Timesteps} & \makecell{F1 \\ Score} \\
\toprule[1.5pt]
    \checkmark & & & & 0.646 \\
    & \checkmark & &  & 0.653 \\
    & & \checkmark &  & 0.664 \\
   & & & \checkmark & 0.649 \\
   \checkmark & \checkmark & \checkmark  & \checkmark & $\bm{0.665}$ \\ 
   \bottomrule[1.5pt]
\end{NiceTabular}
\end{table}
\begin{itemize}[noitemsep,topsep=0pt,leftmargin=*]
\item \textbf{Structured masking strategies perform best}: Table \ref{tab:mask_ablation} shows results from ablating the masking strategies. 
Unlike other masked autoencoder methods \citep{he2022masked}, we found that combining structured masking with random masking outperforms random masking alone.
\item \textbf{Pre-training Presto is critical to achieve strong performance}: In Tables \ref{tab:reg}, \ref{tab:eurosat_finetune} and Table \ref{tab:s2agri}, we compared the performance of a \hlc[orange!20]{randomly-initialized} Presto architecture with the \hlc[MidnightBlue!20]{pre-trained model}. Pre-training yielded a significant increase in performance (a 50\% increase in accuracy on the S2-Agri$_{100}$ dataset). Even when the downstream training dataset size was large (EuroSat has 21,600 training samples), pre-training yielded a 14\% increase in accuracy given RGB inputs and up to 22\% increase in accuracy at lower resolutions (Table \ref{tab:eurosat_finetune_full}). For TreeSatAI with S1 data (Table \ref{tab:treesat-results-full}), a randomly initialized model slightly outperformed the pre-trained model. We hypothesize that this is due to the difference in input relative to the pre-training data, since the TreetSatAI input consists of a single image from only one timestep and one channel group.
\item \textbf{Presto's performance scales with model size}: To measure how different model sizes affect Presto's performance, we pre-trained two larger Presto variants: a deeper variant with 4 encoder layers instead of 2, and a wider variant with a doubled encoder size (Table \ref{tab:scaling}). Performance improved as model size increased, suggesting that practitioners who can afford greater computational costs could obtain better results by training a larger Presto model.
\end{itemize}

\section{Discussion \& Conclusion} \label{sec:conclusion}

\paragraph{Limitations} Presto is designed to ingest 10m/px resolution imagery and is pre-trained on products at this scale. This decision is motivated by the free, global availability over time of products at this scale (such as Sentinel-1 and Sentinel-2). Presto does not natively process very-high resolution imagery such as $<1$ m/px imagery from commercial satellites or drones, which can be costly and often lack complete coverage globally and temporally. In addition, Presto is a pixel-timeseries model. While we demonstrated Presto’s flexibility on single-timestep image datasets, image-based models may be preferred if a user's goal is to process entire images to make a prediction. We observed that Presto's performance on the EuroSAT dataset plateaued as the input resolution increased (Table \ref{results:eurosat}), due to images from classes where the relevant pixels for the class are a minority of the pixels in the image (e.g., highways). In such scene classification challenges, image-based models which can learn the shape of the relevant pixels may be better suited. We discuss this further in Section \ref{sec:highways}. 

\paragraph{Conclusion} 
\begin{table}
\centering
\footnotesize
\caption{\textbf{Effect of model size on validation performance}. To understand the effect of model size on performance, we pre-train two larger variants of Presto. As in Table \ref{tab:mask_ablation}, we measure Presto$_{R}$'s performance on the CropHarvest validation task. The number of parameters includes both the encoder and decoder parameters. The FLOPS are computed for a ``full'' input (12 timesteps, with no missing channels), when passed through the encoder and decoder.} \label{tab:scaling}
\vskip 0.15in
\begin{NiceTabular}{rrrrr}
    \CodeBefore
    \rowcolor{MidnightBlue!20}{2}
    \Body
\toprule[1.5pt]
Depth & Width & \makecell{\# params \\(M)} & \makecell{FLOPs \\(M)} & \makecell{F1\\ score} \\
\toprule[1.5pt]
2 & 128 & 0.81 & 88.94 & 0.665 \\
2 & 256 & 2.02 & 220.81 & 0.687 \\
4 & 128 & 1.21 & 132.42 & 0.669 \\
\bottomrule[1.5pt]
\end{NiceTabular}
\end{table}
We present Presto: a lightweight, pre-trained timeseries transformer for remote sensing. By leveraging structure unique to remote sensing data---specifically, (i) an important temporal dimension, (ii) associated metadata and (iii) a diversity of sensors, we are able to train an extremely lightweight model which achieves state-of-the-art results in a wide variety of globally distributed evaluation tasks. Computational efficiency is of paramount importance in remote sensing settings and often determines which models ultimately get selected for deployment. We demonstrated that strong performance can be achieved while meeting this constraint, and that self-supervised learning can provide significant benefits even for small models.

\subsection*{Impact statement}
Machine learning applications to remote sensing have a wide range of societally beneficial outcomes, ranging from tracking progress on sustainable development goals \citep{ferreira2020monitoring} to improved weather forecasting \citep{english2013impact, voosen2020europe} to disaster management \citep{kansakar2016review}. 

Presto is designed to be accessible to a wide range of practitioners; we achieve this by only training Presto on publicly available data and by keeping the model size small enough so it can be leveraged in compute-constrained environments. In addition to increasing Presto's accessibility, its small size also lowers its carbon footprint \citep{strubell2019energy}.

As described by \citet{tuia2023artificial}, a natural concern when applying machine learning algorithms to remote sensing data is its use to collect information about individuals who are unaware that data is being collected, and therefore cannot consent to this practice. We therefore encourage deployment of Presto in collaboration with local communities and stakeholders \citep{maui,kshirsagar2021becoming,nakalembe2023considerations}.

\subsection*{Acknowledgements}
This work was supported by NASA under the NASA Harvest Consortium on Food Security and Agriculture (Award \#80NSSC18M0039). This research was enabled in part by compute resources provided by Mila (mila.quebec); in addition, we acknowledge material support from NVIDIA Corporation in the form of computational resources. We thank Esther Rolf and Caleb Robinson for reviewing drafts of this manuscript. 

\bibliography{bib}

\appendix
\onecolumn
\section{Appendix}

\subsection*{Reproducibility}
All code and data used to train and evaluate Presto will be made available upon publication, and the code is currently available at \url{https://github.com/nasaharvest/presto}. In addition, we discuss specific implementation details in Appendices \ref{app:pretraining} and \ref{app:downstream}. We have strived to make the Presto codebase accessible to other practitioners; to this end, we include a demo Jupyter notebook demonstrating how Presto can be applied to a new downstream task, which is available at \url{https://github.com/nasaharvest/presto/blob/main/downstream_task_demo.ipynb}.

\subsection{Pre-training details}\label{app:pretraining}
We outline training hyperparameters below:
\begin{itemize}[noitemsep,topsep=0pt,leftmargin=*]
\item \textbf{Training length}: We train the model for 20 epochs, with a batch size of $4096$ (resulting in $5950$ batches per epoch). On a single NVIDIA V100 GPU, this takes 43 $\frac{1}{4}$ hours.
\item \textbf{Optimizer and learning rate}: We train the model with an AdamW optimizer. We use a cosine annealing schedule for our learning rate, with a maximum learning rate of 0.001 at the $2^\text{nd}$ epoch. We apply a weight decay of 0.05, and $\beta$s of (0.9, 0.95). 
\item \textbf{Masking}: We use a masking ratio of $0.75$, randomly selecting (for each instance) a masking strategy from the ones described in Section \ref{sec:masking}. If the masking strategy cannot mask the right number of tokens, we randomly mask additional tokens to achieve the correct masking ratio.
\end{itemize}

\subsubsection{Pre-training data} \label{app:data}

\begin{figure}[ht!]
    \centering
    \includegraphics[width=0.8\textwidth]{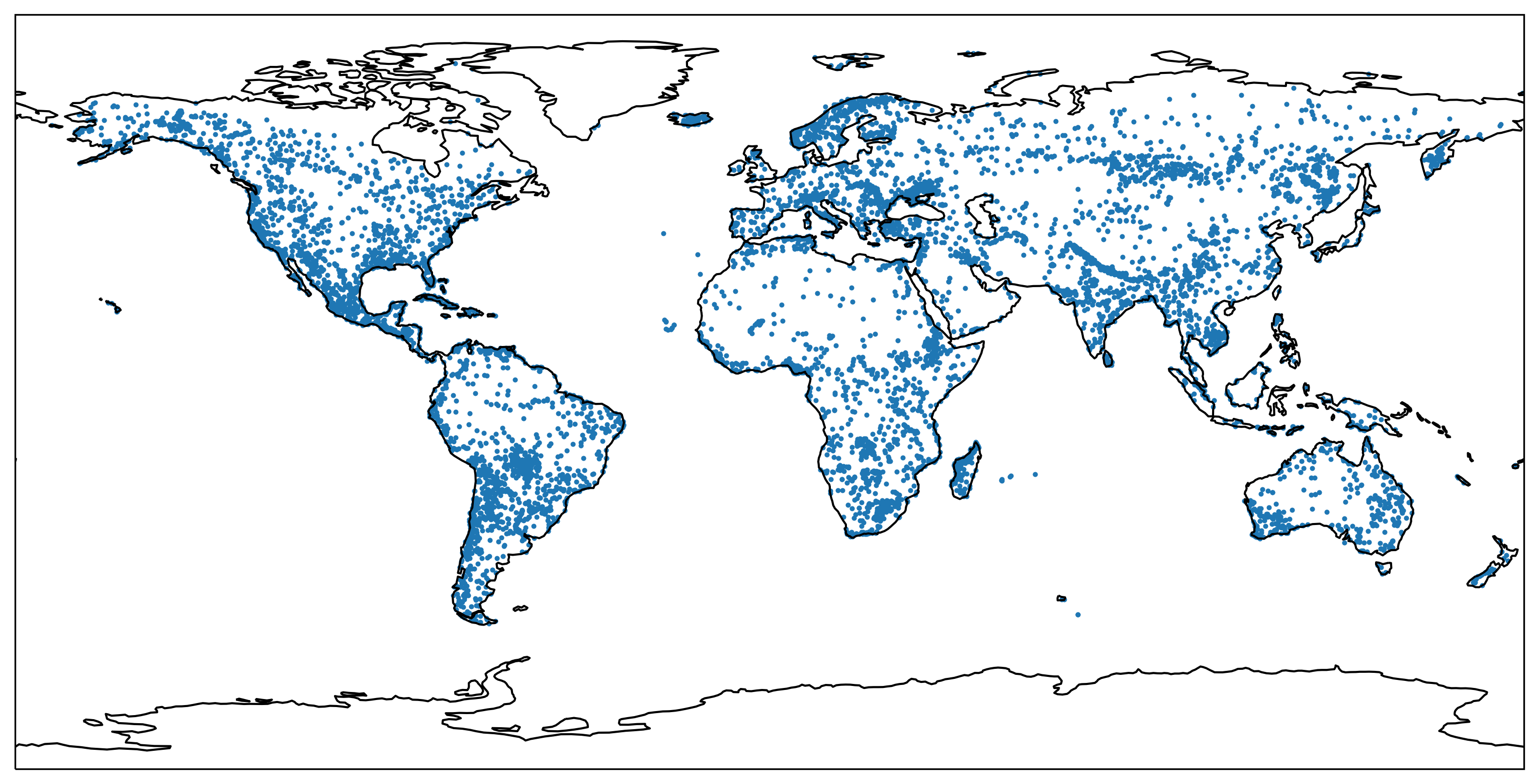}
    \caption{The distribution of the pre-training dataset described in Section \ref{sec:data}.}
    \label{fig:dynamic_world}
\end{figure}

Remote sensing models can be deployed in a wide range of geographies, with few labelled datapoints available at fine-tuning time \citep{rapidresponse,bohm2022sar}. We therefore aim to collect a globally representative pre-training dataset.
We achieve this by following the sampling strategy used by Dynamic World \citep{DynamicWorld}. We divide the Earth into three regions: the Western Hemisphere and two regions in the Eastern Hemisphere. These regions are further divided into ecoregions, and stratified samples are gathered from each region using land cover classes as sampling strata. 
Figure \ref{fig:dynamic_world} shows the resulting geographical distribution.
Each sample represents a $510\times510$ pixel tile with a spatial resolution of 10 meter per pixel. 
To obtain pixel-timeseries we grid-sample 2,500 pixels from each sample, yielding a total of 21,535,000 pixel samples (each with 24 one-month timesteps).

\subsubsection{Input data}\label{app:input_data}
We leverage the following data products when pre-training Presto:
\begin{itemize}[noitemsep,topsep=0pt,leftmargin=*]
    \item \textbf{Sentinel-1 Synthetic Aperture Radar observations} (S1): The VV (emit and receive at vertical polarization) and VH (emit at vertical and receive at horizontal polarization) bands: 2 real-valued dynamic values per monthly timestep.
    \item \textbf{Sentinel-2 Multispectral images} (S2): We removed the 60m resolution bands, yielding bands with 10m and 20m resolution with channels in the visible, near-infrared and short-wave infrared range: 10 real-valued dynamic values per timestep.
    \item \textbf{ERA5 Climate Reanalysis Meteorological data} (ERA5): Monthly total precipitation and temperature at 2 metres above the ground: 2 real-valued dynamic values per timestep.
    \item \textbf{NDVI} \citep{rouse1974monitoring}: Computed from the red (B4) and near-infrared (B8) Sentinel-2 bands: 1 real-valued dynamic value per timestep.
    \item \textbf{Dynamic World Land Cover classes} \citep[DW, ][]{DynamicWorld}: Land cover classes produced for every non-cloudy Sentinel-2 image: 1 dynamic categorical value from the set of possible classes $\mathcal{V}$ per timestep. We took the mode of classes for all timesteps within a month.
    \item \textbf{Topography data} (TG), from the Shuttle Radar Topography Mission's Digital Elevation Model: The elevation and slope of each pixel, real-valued and static in time.
    \item \textbf{Coordinates} (Loc): 3D static in time Cartesian coordinates computed from the latitude and longitude of the pixel's geographical location: $s_{\text{Loc}} = [\cos(\llat) \times \cos(\llon), \cos(\llat) \times \sin(\llon), \sin(\llat)]$.
\end{itemize}

\subsubsection{Channel Groups} \label{sec:channel_groups}
As described in Section \ref{sec:tokenization}, we transform the pixel timeseries $x$ into a number of tokens, where each token is a linear transformation of a subset of the input channels. We group together channels which (i) come from the same sensor or product, (ii) have equivalent native spatial resolutions and (iii) represent similar parts of the electromagnetic spectrum (for Sentinel-2 channel groups). We group the input data into the following channel groups:

\begin{itemize}[noitemsep,topsep=0pt,leftmargin=*]
\item \textbf{Sentinel-1}: The VV and VH bands from the Sentinel-1 sensor
\item \textbf{Sentinel-2 RGB}: The B2, B3 and B4 bands from the Sentinel-2 sensor
\item \textbf{Sentinel-2 Red Edge}: The B5, B6 and B7 bands from the Sentinel-2 sensor
\item \textbf{Sentinel-2 Near Infra Red (10m)}: The B8 band from the Sentinel-2 sensor
\item \textbf{Sentinel-2 Near Infra Red (20m)}: The B8A band from the Sentinel-2 sensor
\item \textbf{Sentinel-2 Short Wave Infra Red}: The B11 and B12 bands from the Sentinel-2 sensor
\item \textbf{NDVI}: The normalized difference vegetation index, calculated from the Sentinel-2 B4 and B8 bands.
\item \textbf{ERA5 Climatology}: Precipitation and temperature at 2m from the ERA5 Climate Reanalysis product
\item \textbf{Topography}: The elevation and slope of a pixel, calculated by the SRTM's DEM
\item \textbf{Location}: The cartesian coordinates of a pixel, computed from the pixel's latitude and longitude
\end{itemize}

\subsection{FLOP calculations}

\begin{table}
\centering
\caption{Model sizes and FLOPs required to encode a single EuroSat image (or pixel, for Presto), as measured by the \texttt{thop} library. When plotting results in Table \ref{results:eurosat}, we multiply the FLOPs for Presto by the number of pixels encoded for an image. At its highest resolution, EuroSAT images are $64 \times 64$, so Presto FLOPs for a full resolution image can be obtained by multiplying the per-pixel FLOPs by 4,096. We include this value in brackets for completeness.}
\begin{tabular}{llrr}
\toprule[1.5pt]
Model & Backbone & Params (M) & MegaFlops \\
\toprule[1.5pt]
SatMAE (RGB) \citep{cong2022satmae} & ViT-Large & 303.10 & 59,685.69 \\
SatMAE (MS) \citep{cong2022satmae} & ViT-Large & 305.96 & 535,515.25 \\
ScaleMAE \citep{reed2022scale} & ViT-Large & 303.10 & 59,685.69 \\
ConvMAE \citep{gao2022convmae} & ConvMAE-Large & 88.78 & 23,315.58 \\
SeCo \citep{manas2021seasonal} & ResNet-18 & 11.69 & 149.37 \\
GASSL \citep{ayush2021geography} & ResNet-18 & 11.69 & 149.37 \\
Presto RGB pixel (image) & Presto & 0.40 & 0.79 (3,235.84) \\
Presto MS pixel (image) & Presto & 0.40 & 2.37 (9,707.52) \\
\bottomrule[1.5pt]
\end{tabular}
    \label{tab:flops}
\end{table}

We use the \texttt{thop} library (\url{https://github.com/Lyken17/pytorch-OpCounter}) to calculate the FLOPs required to encode a EuroSAT image (as plotted in Table \ref{results:eurosat}(b)). For the SatMAE, ScaleMAE and ConvMAE models, all images were resized to $224 \times 224$, so the FLOPs required to encode an image is independent of resolution. For Presto, we computed the FLOPs required to encode a single pixel and multiplied this by the number of pixels in an image at each resolution (e.g. the ``64'' resolution has $64 \times 64$ pixels, so we multiply the FLOPs required to encode a single pixel by $64 \times 64 = 4096$). The FLOPs calculated by the \texttt{thop} library are recorded in Table \ref{tab:flops}.

\subsection{Baselines} \label{app:baselines}

In addition to task-specific baselines, we benchmark Presto against:
\begin{itemize}[noitemsep,topsep=0pt,leftmargin=*]
\item \textbf{Random Forests}: Random forests are powerful baselines in remote sensing as they they remain competitive with state-of-the-art methods \citep{Pelletier2019Temporal,rapidresponse}. Tree-based methods, especially random forests, are commonly deployed in large-scale machine learning for remote sensing applications \citep{hansen2013high,worldcerealsbenchmark,di2022annual}.
\item \textbf{MOSAIKS-1D}: We adapt MOSAIKS \citep{rolf2021generalizable} for timeseries data. MOSAIKS-1D uses patches from the pre-training dataset and convolves over the temporal dimension instead of the spatial dimension. We benchmark MOSAIKS-1D on all timeseries evaluation tasks. Because this does not work for categorical inputs, we exclude Dynamic World. As with Presto, we use the output features with random forests (MOSAIKS-1D$_{RF}$) and with regressions (MOSAIKS-1D$_{R}$).
\end{itemize}

\begin{table*}
    \centering
    \caption{Full results for regression tasks from Table \ref{tab:reg}, including standard error computed from three runs.}\label{tab:reg_se}
    \begin{tabular}{lrrr}
    \toprule[1.5pt]
    & Fuel Moisture & Algae Blooms & Mean difference \\
    \toprule[1.5pt]
    Linear Regression & 28.20 & \bm{$0.850$} & 0\% \\
    Random Forest & {\color{blue} \bm{$23.84 \pm 0.42$}} & $1.249 \pm 0.02$ & 15.7\%\\
    MOSAIKS-1D$_{RF}$ & $28.75 \pm 0.15$ & $0.972 \pm 0.01$ & 8.15\%\\
    Presto$_{FT}$ (random init.) & $26.07 \pm 0.52 $ & $0.955 \pm 0.05$ & 2.40\% \\
    \midrule[0.5pt]
    Presto$_{FT}$ & \bm{$25.28 \pm 0.30$}  & {\color{blue}\bm{$0.815 \pm 0.03$}} & ${\color{blue}\bm{-7.24\%}}$ \\
    Presto$_{RF}$ & $25.98 \pm 0.66$ & $0.884 \pm 0.01$ & \bm{$-1.94\%$} \\
    \bottomrule[1.5pt]
    \end{tabular}
\end{table*}

\subsection{Downstream Results} \label{app:downstream}

We include complete results for the evaluation tasks. These include error bars, as well as additional results reported for the CropHarvest (Table \ref{tab:cropharvest-results-full} and Figure \ref{fig:cropharvest_over_time}), regression tasks (Table \ref{tab:reg_se}), EuroSAT (Tables  \ref{tab:eurosat_finetune_full}, \ref{tab:eurosat-results-full} and \ref{tab:eurosat-results-extended}), TreeSatAI (Table \ref{tab:treesat-results-full}) and Sen2-Agri$_{100}$ (Table \ref{tab:s2agri-full}) datasets.

We run all downstream classifiers with 3 seeds ($0, 42, 84$), with the exception of the kNN classifiers and the linear regression (which are deterministic). In the tables in the main paper (Tables \ref{tab:cropharvest-results}, \ref{tab:treesat}, \ref{tab:s2agri} and \ref{tab:reg}) we report the average of these runs; the standard error is reported in Tables \ref{tab:cropharvest-results-full},\ref{tab:treesat-results-full}, \ref{tab:s2agri-full} and \ref{tab:reg_se}.

\begin{itemize}[noitemsep,topsep=0pt,leftmargin=*]
\item \textbf{Presto as a feature extractor}: When used as a feature extractor, a random forest, regression of K-nearest-neighbours classifier is trained on Presto's output embeddings. In this case, we use scikit-learn models with the default hyperparameters. For the CropHarvest tasks, the class labels are extremely imbalanced; we therefore set \texttt{class\_weight} equal to \texttt{balanced} for those tasks, for both Presto and MOSAIKS-1D.
\item \textbf{Fine-tuning Presto}: When fine-tuning Presto, we use the same hyperparameters across all tasks: an AdamW optimizer with a learning rate of \texttt{3e-4} and weight decay of $0.05$.
\end{itemize}

As discussed in Section \ref{sec:image}, we obtain per-image predictions using Presto by computing a mean and standard deviation of Presto's output pixels, and passing a concatenation of these two vectors to a downstream classifier. This is illustrated in Figure \ref{im:aggregates}.

\subsection{Disentangling the effect of pre-training}
To understand the effect of pre-training Presto, we fine-tune Presto and train it from scratch on EuroSat (Table \ref{tab:eurosat_finetune}), the regression tasks (Table \ref{tab:reg} in the main paper) and TreeSatAI (Table \ref{tab:treesat-results-full}). We omit the CropHarvest dataset because it was expressly designed as a few-shot-learning dataset. Its small size makes the construction of validation sets with which to control the finetuning (e.g. with early stopping) challenging.

Overall, we find a consistent and significant improvement from the use of pre-trained Presto compared to a randomly initialized version of the model. For the EuroSat task, pre-training consistently delivers an incresse in accuracy score $> 0.1$ (representing increases in accuracy of up to 25\%). This effect is consistent with what we observe on the TreeSatAI dataset for S2 data and on the regression tasks (where pre-training reduces RMSE by to 15\% on the algae blooms task). For the TreeSatAI dataset with S1 data, pre-training penalizes the model compared to random initialization - we hypothesize that this is due to the difference in input (a single timestep and single channel group image) relative to the pre-training data. The benefit of pre-training effect is especially pronounced on the S2-Agri$_{100}$ dataset; we hypothesize this is due to the small training set size.

\begin{table}
\footnotesize
    \centering
    \caption{Accuracy results for pre-trained and from-scratch Presto when fine-tuned on EuroSat, at varying resolutions. We hypothesize that the drop in performance for the full resolution (64) RGB input is due to the model construction; the model outputs for all pixels in the image (4,096 pixels for the full resolution) are aggregated and passed to a linear layer for classification, yielding a noisy gradient signal.} \label{tab:eurosat_finetune_full}
    \begin{tabular}{llrrrrrr}
    \toprule[1.5pt] \\
    Resolution & & 2 & 4 & 8 & 16 & 32 & 64 \\
    \toprule[1.5pt]
    random init. & \multirow{2}*{RGB} & $0.703 \pm 0.005$ & $0.684 \pm 0.032$ & $0.694 \pm 0.013$ & $0.739 \pm 0.004$ & $0.750 \pm 0.018$ & $0.745 \pm 0.009$ \\
    pre-trained & & $0.792 \pm 0.010$ & $0.837 \pm 0.006$ & $0.847 \pm 0.016$ & $0.865 \pm 0.006$ & $0.872 \pm 0.002$ & $0.849 \pm 0.004$ \\ 
    \midrule
    random init. & \multirow{2}*{MS} & $0.837 \pm 0.014$ & $0.884 \pm 0.010$ & $0.895 \pm 0.006$ & $0.907 \pm 0.13$ & $0.924 \pm 0.005$ & $0.924 \pm 0.003$ \\
    pre-trained & & $0.898 \pm 0.005$ & $0.925 \pm 0.004$ & $0.939 \pm 0.000$ & $0.950 \pm 0.002$ & $0.958 \pm 0.001$ & $0.953 \pm 0.004$ \\ 
    \bottomrule[1.5pt]
    \end{tabular}
\end{table}

\begin{table}
\centering
\caption{Additional results for the CropHarvest task. In addition to the F1 scores reported in the main paper, we report AUC ROC scores, with standard error bars computed with three runs.}
  \begin{tabular}{llrrrr}
    \toprule[1.5pt]
     & Model & Kenya & Brazil & Togo & Mean \\
     \toprule[1.5pt]
     \multirow{5}*{F1} & Random Forest & $0.559 \pm 0.003$ & $0.000 \pm 0.000$ & $0.756 \pm 0.002$ & 0.441 \\
     & MOSAIKS-1D$_{R}$ & $0.790 \pm 0.027$ & $0.746 \pm 0.084$ & $0.679 \pm 0.024$ & 0.738 \\
      & TIML & $0.838 \pm 0.000$ & $0.835 \pm 0.012$ & $0.732 \pm 0.002$ & $0.802$ \\
      \cmidrule[0.5pt]{2-6}
      & Presto$_{R}$ & $0.816 \pm 0.000$ & \bm{$0.891 \pm 0.000$} & \bm{$0.798 \pm 0.000$} & $0.835$ \\
      & \hspace{3mm} no DW & $\bm{0.861 \pm 0.000}$ & $0.888 \pm 0.000$ & $0.760 \pm 0.000$ & \bm{$0.836$} \\
      \midrule
          \multirow{5}*{AUC ROC} & Random Forest & $0.578 \pm 0.006$ & $0.941 \pm 0.004$ & $0.892 \pm 0.001$ & 0.803 \\
     & MOSAIKS-1D$_{R}$ & $0.693 \pm 0.036$ & $0.890 \pm 0.038$ & $0.836 \pm 0.005$ & 0.806 \\
      & TIML & $0.794 \pm 0.003$ & $0.988 \pm 0.001$ & $0.890 \pm 0.000$ & $0.890$ \\
        \cmidrule[0.5pt]{2-6}
      & Presto$_{R}$ & $0.834 \pm 0.000$ & \bm{$0.997 \pm 0.000$} & \bm{$0.921 \pm 0.000$} & $0.917$ \\
      & \hspace{3mm} no DW & $\bm{0.863 \pm 0.000}$ & $0.989 \pm 0.000$ & $0.912 \pm 0.000$ & \bm{$0.921$} \\ 
    \bottomrule[1.5pt]
  \end{tabular}
    \label{tab:cropharvest-results-full}
\end{table}

\begin{table}
\centering
\caption{Additional results for the EuroSat task - results for the ScaleMAE, SatMAE and ConvMAE models are from \citep{reed2022scale}. We report kNN classifier results for different values of $k$, and at varying input resolutions.}
  \begin{tabular}[width=\linewidth]{@{}lrrr|rrr|rrr@{}}
    \toprule[1.5pt]
    Resolution & \multicolumn{3}{c}{16} & \multicolumn{3}{c}{32} & \multicolumn{3}{c}{64} \\
    $k$ & $5$ & $20$ & $100$ & $5$ & $20$ & $100$ & $5$ & $20$ & $100$ \\ 
    \toprule[1.5pt]
    SatMAE & 0.729 & 0.727 & 0.695 & 0.871 & 0.876 & 0.854 & 0.934 & 0.931 & 0.913 \\
    ScaleMAE & 0.751 & 0.744 & 0.699 & 0.912 & 0.901 & 0.869 & \textbf{0.960} & 0.956 & 0.935 \\
    ConvMAE & 0.835 & 0.826 & 0.788 & 0.909 & 0.898 & 0.863 & 0.947 & 0.940 & 0.914 \\
    \midrule[0.5pt]
    Presto (RGB) & 0.869 & 0.828 & 0.713 & 0.869 & 0.829 & 0.712 & 0.869 & 0.829 & 0.713 \\
    Presto (MS) & \textbf{0.916} & 0.892 & 0.844 & \textbf{0.920} & 0.892 & 0.846 & 0.921 & 0.893 & 0.846 \\
    \bottomrule[1.5pt]
    \end{tabular}
    \label{tab:eurosat-results-full}
\end{table}

\begin{table}
\centering
\caption{Additional results for the EuroSat task for Presto when run with reduced resolutions (compared to those used by \citep{reed2022scale} and reported in Table \ref{tab:eurosat-results-full}). We report kNN classifier results for different values of $k$, and at varying input resolutions.}
    \label{tab:eurosat-results-extended}
  \begin{tabular}[width=\linewidth]{@{}lrrr|rrr|rrr@{}}
    \toprule[1.5pt]
    Resolution & \multicolumn{3}{c}{2} & \multicolumn{3}{c}{4} & \multicolumn{3}{c}{8} \\
    $k$ & $5$ & $20$ & $100$ & $5$ & $20$ & $100$ & $5$ & $20$ & $100$ \\ 
    \toprule[1.5pt]
    Presto (RGB) & 0.843 & 0.811 & 0.699 & 0.860 & 0.820 & 0.706 & 0.869 & 0.826 & 0.710 \\
    Presto (MS) & 0.873 & 0.852 & 0.799 & 0.895 & 0.874 & 0.824 & 0.911 & 0.886 & 0.838 \\
    \bottomrule[1.5pt]
    \end{tabular}
\end{table}

\begin{table}
\centering
\footnotesize
\caption{Additional results for the TreeSatAI (as in \citep{ahlswede2023treesatai}, we report precision and recall in addition to $F_1$ score and mAP). In addition, we report the results of finetuning Presto (Presto$_{FT}$) from the pre-trained weights and from a random initialization.}\label{tab:treesat-results-full}
  \begin{tabular}[width=\linewidth]{@{}lll|rrrr@{}}
    \toprule[1.5pt]
    Model & Data & Aggregation & $F_1$ & mAP & Precision & Recall \\ 
    \toprule[1.5pt]
    MLP & \multirow{10}*{S1} & \multirow{5}*{Weighted} & $10.09$ & $29.42$ & $33.29$ & $7.13$ \\
    LightGBM & & & $11.86$ & $32.79$ & $37.96$ & $8.06$ \\
    Presto$_{FT}$ (random init.) & & & $40.36 \pm 0.77$ & $39.77 \pm 0.79$ & $30.69 \pm 0.82$ & $64.69 \pm 1.09$ \\
    Presto$_{FT}$  & & & $38.69 \pm 0.78$ & $37.41 \pm 0.58$ & $30.09 \pm 0.74$ & $61.20 \pm 0.85$ \\
    Presto$_{RF}$ & & & $38.34 \pm 0.07$ & $35.45 \pm 0.03$ & $29.67 \pm 0.07$ &  $57.23 \pm 0.06$ \\
    \cmidrule{3-7}
    MLP & & \multirow{5}*{Micro} & $12.82$ & $33.09$ & $63.01$ & $7.13$ \\
    LightGBM & & & 14.07 & 35.11 & $55.49$ & $8.06$  \\
    Presto$_{FT}$ (random init.) & & & $42.04 \pm 0.73$ & $43.00 \pm 0.80$ & $31.20 \pm 1.00$ & $64.69 \pm 1.09$ \\
    Presto$_{FT}$ & & & $41.65 \pm 0.46$ & $40.75 \pm 0.69$ & $31.58 \pm 0.47$ & $61.20 \pm 0.85$ \\
    Presto$_{RF}$ & & & $40.79 \pm 0.04$ & $38.64 \pm 0.02$ & $31.69 \pm 0.03$ & $57.23 \pm 0.06$ \\
    \midrule
    MLP & \multirow{10}*{S2} & \multirow{5}*{Weighted} & $51.97$ & $64.19$ & $74.59$ & $42.23$ \\
    LightGBM & & & $48.17$ & $61.99$ & $74.27$ & $40.04$ \\
     Presto$_{FT}$ (random init.) & & & $52.74 \pm 0.50$ & $57.24 \pm 0.64$ & $45.87 \pm 1.17$ & $64.29 \pm 1.51$ \\
    Presto$_{FT}$ & & & $53.63 \pm 0.42$ & $59.16 \pm 1.24$ & $47.15 \pm 1.40$ & $65.11 \pm 3.21$ \\
    Presto$_{RF}$ & & & $55.29 \pm 0.08$ & $61.53 \pm 0.09$ & $56.93 \pm 0.07$ & $58.56 \pm 0.09$ \\
    \cmidrule{3-7}
    MLP & & \multirow{5}*{Micro} & $54.49$ & $65.83$ & $77.18$ & $42.23$ \\
    LightGBM & & & $52.52$ & $61.66$ & $76.27$ & $40.04$ \\
    Presto$_{FT}$ (random init.) & & & $52.56 \pm 0.41$ & $58.08 \pm 0.66$ & $44.56 \pm 1.03$ & $64.29 \pm 1.51$ \\
    Presto$_{FT}$ & & & $53.31 \pm 0.18$ & $59.77 \pm 1.13$ & $45.51 \pm 1.46$ & $65.11 \pm 3.21$\\
    Presto$_{RF}$ & & & $58.29 \pm 0.06$ & $63.31 \pm 0.06$ & $58.04 \pm 0.05$ & $58.56 \pm 0.09$ \\
    \bottomrule[1.5pt]
    \end{tabular}
\end{table}

\begin{table}
\centering
\caption{Full results on the S2-Agri$_{100}$ dataset, including standard errors obtained from 3 runs. To obtain standard errors for the SITS-Former, we run the official code (\url{https://github.com/linlei1214/SITS-Former}) with 3 seeds. Best results are \textbf{highlighted}.} \label{tab:s2agri-full}
\vskip 0.15in
\begin{NiceTabular}{lccrrrr}
\toprule[1.5pt]
     & Params (M) & Pre-trained? & OA & $\kappa$ &$F_{1}$ \\
    \toprule[1.5pt]
    \multirow{2}{*}{\makecell{SITS \\ Former}} & \multirow{2}{*}{2.5} & & $65.13 \pm 3.01$ & $0.55 \pm 0.03$ & $42.12 \pm 0.52$ \\
    & & \checkmark & $67.03 \pm 2.24$ & $0.56\pm 0.02$ & $\bm{42.83} \pm 0.30$ \\
    \midrule
    \multirow{2}{*}{Presto} & \multirow{2}{*}{0.4} & & $45.98 \pm 2.74$ & $0.35 \pm 0.02$ & $27.45 \pm 0.64$\\
    & & \checkmark & $\bm{68.89} \pm 1.05$ & $\bm{0.58} \pm 0.01$ & $40.41 \pm 0.25$\\
    \bottomrule[1.5pt]
\end{NiceTabular}
\end{table}

\subsection{Presto's failure modes} \label{sec:highways}

\begin{figure}[ht!]
\begin{center}
\includegraphics[width=0.5\linewidth]{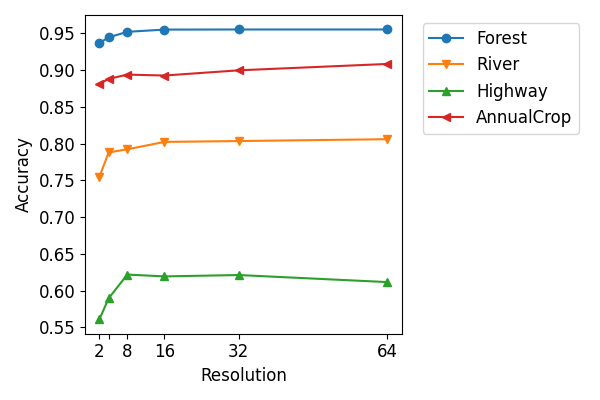}
\caption{Accuracy of kNN@5 classifier with Presto RGB representations on the EuroSat dataset vs. the input resolution, for different categories. Some categories have been left out for clarity.}
\label{fig:eurosat_per_category}
\end{center}
\end{figure}

\begin{figure}[ht!]
\begin{center}
\centering
\begin{subfigure}{0.25\textwidth}
 \centering
 \includegraphics[width=\textwidth]{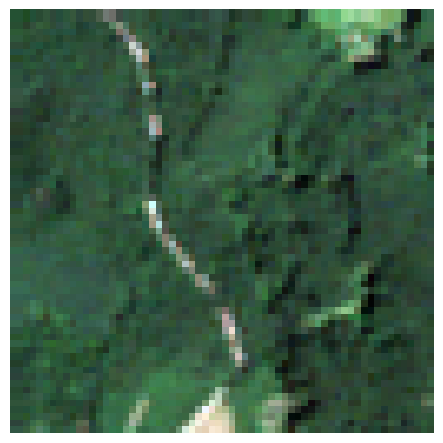}
 \caption{Forest}
 \label{fig:eurosat_forest}
\end{subfigure}\begin{subfigure}{0.25\textwidth}
 \centering
 \includegraphics[width=\textwidth]{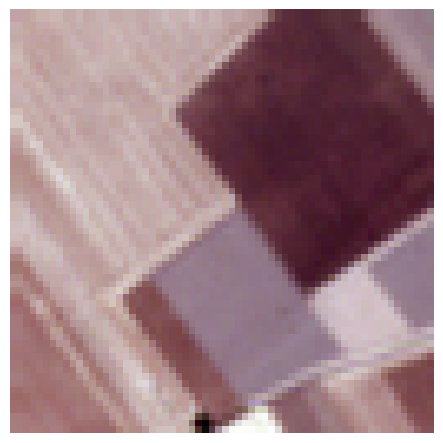}
 \caption{Annual Crop}
 \label{fig:eurosat_pasture}
\end{subfigure}\begin{subfigure}{0.25\textwidth}
 \centering
 \includegraphics[width=\textwidth]{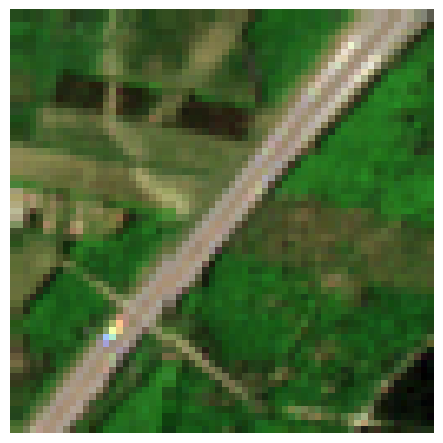}
 \caption{Highway}
 \label{fig:eurosat_highway}
\end{subfigure}\begin{subfigure}{0.25\textwidth}
 \centering
 \includegraphics[width=\textwidth]{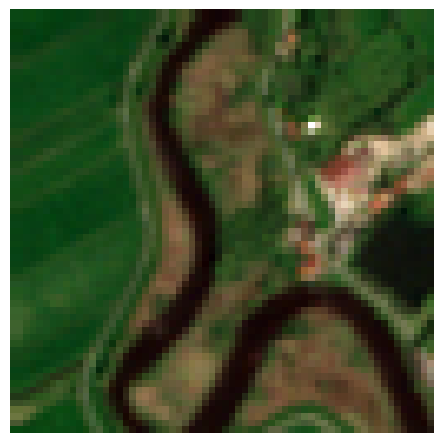}
 \caption{River}
 \label{fig:eurosat_river}
\end{subfigure}
\caption{the RGB bands of example images from EuroSat classes.}
\label{fig:eurosat_examples}
\end{center}
\end{figure}

Presto processes pixel-timeseries independently, without spatial context from other pixels or locations.
This means that when we make image-based predictions (such as for scene classification), Presto's independent pixel representations must be aggregated into a single prediction.
We opt for a simple concatenation of the element-wise mean and standard deviation of the representations, from which a classifier makes a prediction. Information gets lost in such a simple aggregation, which impacts Presto's performance on such tasks.

For example, Presto's performance on the EuroSat dataset reaches a plateau when increasing the input resolution. 
As Figure \ref{fig:eurosat_per_category} shows, this is mainly caused by a failure to accurately predict specific classes (for example, the \textit{Highway} and \textit{River} classes).
Figure \ref{fig:eurosat_examples} shows example images for these classes, as well as for the \textit{Forest} and \textit{AnnualCrop} classes, on which Presto achieves higher accuracies.
While in the \textit{Forest} and \textit{AnnualCrop} images, most pixels of the image actually represent the labelled class, in the \textit{Highway} and \textit{River} images only a relatively small part of the image actually contains the label (a highway or river). We hypothesize that since many pixels in the \textit{Highway} and \textit{River} images do not actually represent that class, the crude token-aggregation method we use to represent images is insufficiently discriminative to accurately classify these images.

Other pre-trained remote sensing models use much more powerful mechanisms for aggregating spatial information. For example, ViT models convolve over patches and then apply an attention mechanism between spatial patches. If image-based predictions are needed and these predictions are highly dependent on the occurrence of objects in subregions of the image, models which natively process this important spatial information may be better suited.

We plan on exploring techniques to mitigate this difficulty with Presto in future work.

\end{document}